\pdfoutput=1

\documentclass[11pt]{article}

\usepackage[]{naacl2021}

\usepackage{times}
\usepackage{latexsym}

\usepackage{makecell}
\usepackage{graphicx}
\usepackage{tikz}
\usepackage{multirow}
\usepackage{booktabs,tabularx,enumitem,ragged2e}
\usepackage{url}
\usepackage{amsmath}
\usepackage{dirtytalk}

\usepackage[T1]{fontenc}

\usepackage[utf8]{inputenc}

\usepackage{microtype}

\def\modelname{\textsc{DuCo-StoryGAN}}

\usepackage[normalem]{ulem}

\title{Improving Generation and Evaluation of Visual Stories \\ via Semantic Consistency}

\author{Adyasha Maharana \,\,\,\,\,\,\,\,\,\,\,\,Darryl Hannan\,\,\,\,\,\,\,\,\,\,\,\,\,Mohit Bansal \\
    Department of Computer Science \\
  University of North Carolina at Chapel Hill \\
  {\tt \{adyasha, dhannan, mbansal\}@cs.unc.edu} \\
}

\begin{document}
\maketitle
\begin{abstract}
Story visualization is an underexplored task that falls at the intersection of many important research directions in both computer vision and natural language processing. In this task, given a series of natural language captions which compose a story, an agent must generate a sequence of images that correspond to the captions. Prior work has introduced recurrent generative models which outperform text-to-image synthesis models on this task. However, there is room for improvement of generated images in terms of visual quality, coherence and relevance. We present a number of improvements to prior modeling approaches, including (1) the addition of a dual learning framework that utilizes video captioning to reinforce the semantic alignment between the story and generated images, (2) a copy-transform mechanism for sequentially-consistent story visualization, and (3) MART-based transformers to model complex interactions between frames. We present ablation studies to demonstrate the effect of each of these techniques on the generative power of the model for both individual images as well as the entire narrative. Furthermore, due to the complexity and generative nature of the task, standard evaluation metrics do not accurately reflect performance. Therefore, we also provide an exploration of evaluation metrics for the model, focused on aspects of the generated frames such as the presence/quality of generated characters, the relevance to captions, and the diversity of the generated images. We also present correlation experiments of our proposed automated metrics with human evaluations.\footnote{Code and data: \url{https://github.com/adymaharana/StoryViz}.}

\end{abstract}

\begin{figure}[t]
\centering
\includegraphics[width=0.50\textwidth]{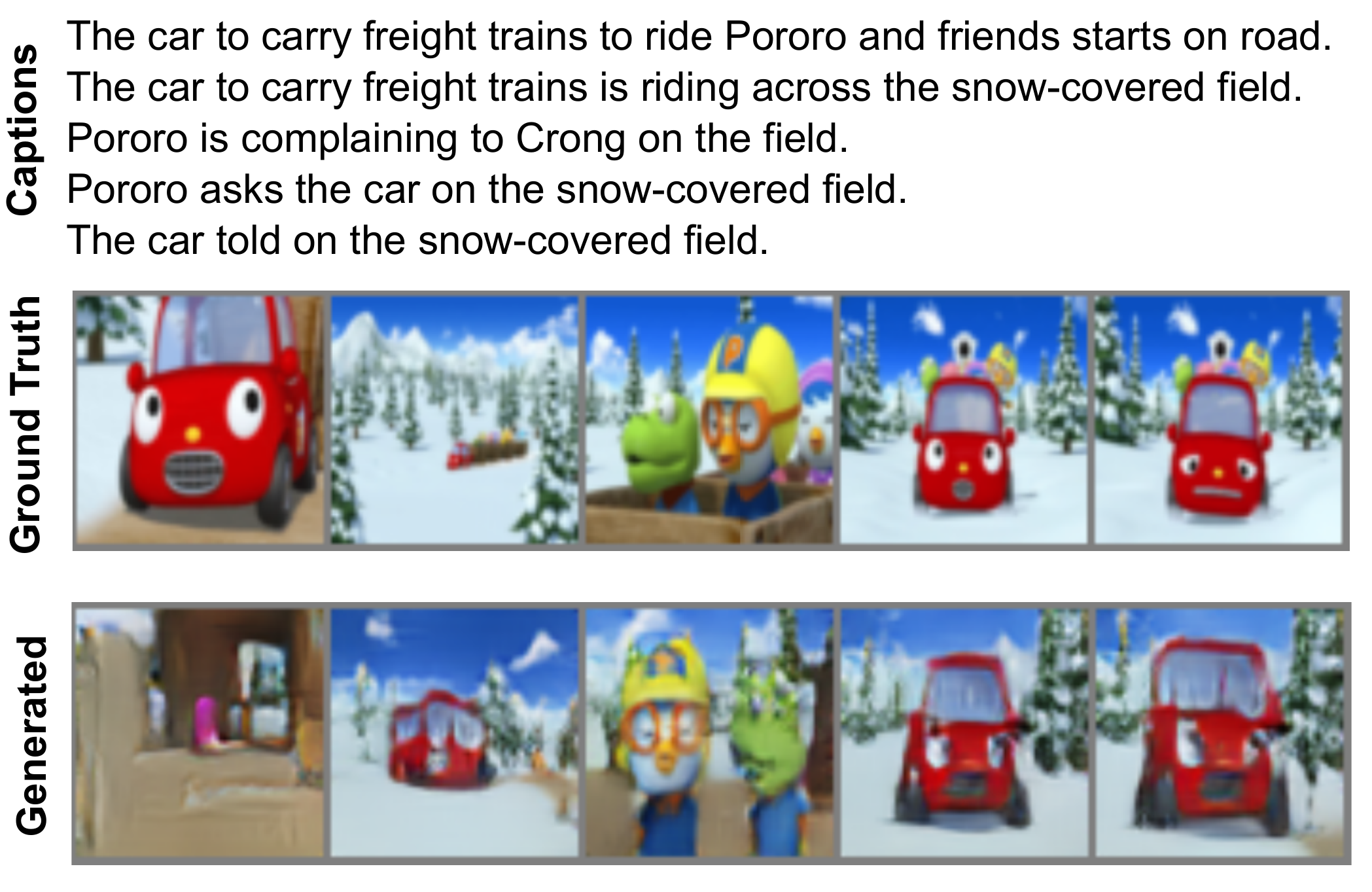}
\caption{Illustration of the Pororo-SV dataset (Captions \& Ground Truth) and the corresponding images generated from our model (Generated).
\label{fig:pororo_example}
\vspace{-15pt}}
\end{figure}

\section{Introduction}
\label{sec:intro}

While generative adversarial networks (GANs) have achieved impressive results on a variety of image generation tasks \cite{dmgan, qiao2019mirrorgan}, the task of story visualization \cite{li2019storygan} is a variation of image generation that is more challenging and underexplored. In this setting, there is a story which consists of a sequence of images along with captions describing the content of the images, e.g., a web comic. The goal of the task is to reproduce the images given the captions (Figure \ref{fig:pororo_example}). The benefits of investigating this task are far reaching. It combines two interesting and challenging sub-areas: text-to-image synthesis and narrative understanding, providing an excellent test bed for exploring and developing multimodal modeling techniques. From an application perspective, such a system could be used to enhance existing textual narratives with visual scenes. This tool would be especially useful to comic artists, who are infamously overworked, allowing them to automatically generate initial drawings speeding up their workflow. Additionally, such a system would have many applications in an educational setting, allowing educators to cater to a more diverse set of learning styles by automatically generating visualizations for a given topic, such as the water cycle in a science lesson. Furthermore, the data in this domain is cartoon-style, meaning the generated images avoid many of the ethical issues associated with real-world data. For a more detailed discussion, see Section \ref{sec:ethics}. 

The challenge of this task extends beyond tasks such as text-to-image or text-to-video synthesis. Namely, there is an explicit, narrative component to the data, which must first be accurately extracted from the text, and then consistently reproduced throughout the images. If the setting or a description of a character is provided in the first caption, this must be carried throughout the scene unless modified by a subsequent caption. Furthermore, the scenes in a single story can change drastically as the story progresses, requiring models to produce a greater variety of images than in a text-to-video task, which typically consists of short videos displaying a single action. To address these issues, we consider the task as proposed in \citet{li2019storygan}, which provides a baseline architecture, StoryGAN, along with datasets for the task. We introduce techniques that build on existing work and are focused on improving consistency across frames, resulting in images of higher visual quality.

First, we augment the model with Dual Learning via video redescription. The output images are fed through a video captioning model, which is trained to reproduce the ground truth story captions. This provides an additional learning signal to the model, forcing it to semantically align with the given narrative. Next, we add a Copy-Transform module that can take generated images from previous timesteps and copy the most relevant features of those images into the next generated frame, thus making the images more consistent in appearance. Finally, we propose the use of Memory-Augmented Recurrent Transformer (MART) \cite{lei2020mart} to model the correlation between word phrases in the input text and corresponding regions in the generated image. The recurrent nature of MART allows for the learning of sophisticated interactions between the image frames, yielding images that are more consistent in terms of character appearances and background imagery. We call the model architecture with the aforementioned additions \textsc{Du(al)}-\textsc{Co(py)}-\textsc{StoryGAN} or \modelname.

Next, we focus on exploring alternative evaluation methods for story visualization models. While modeling improvements are crucial for progressing in this domain, evaluating these models is a challenge in itself. Like many other generative tasks, it is nontrivial to evaluate a story visualization model. Human evaluation is the most reliable option, but its monetary and time costs make this ill-suited to be the only evaluation method. Most prior work relies upon standard GAN evaluation metrics, which may provide some insight into how well the images were reproduced, yet miss out on other aspects of the story visualization task, such as the visual consistency of the setting across frames and global semantic alignment. Therefore, we make evaluation another focal point of the paper, exploring a variety of automatic evaluation metrics, which capture various aspects of the task, e.g., evaluating the quality of the images, the relevance to the story, the diversity of the generated frames, and the model's ability to accurately represent the characters. We present results from our model and baseline models on all metrics along with qualitative results, demonstrating the improvements from our proposed techniques. Using these metrics, we also provide ablation analyses of our model.

Our main contributions can be summarized as:
\begin{enumerate}
    \item For the story visualization task, we improve the semantic alignment of the generated images with the input story by introducing dual learning via video redescription.
    \item We enable \textit{sequentially-consistent} story visualization with the introduction of a copy-transform mechanism in the GAN framework.
    \item We enhance prior modeling techniques in story visualization with the addition of Memory Augmented Recurrent Transformer, allowing the model to learn more sophisticated interactions between image frames.
    \item We present a diverse set of automatic evaluation metrics that capture important aspects of the task and will provide insights for future work in this domain. We also conduct correlation experiments for these metrics with human evaluation. 

\end{enumerate}

\section{Related Work}
\citet{li2019storygan} introduced the task of story visualization and the StoryGAN architecture for sequential text-to-image generation. There have been a few other works that have attempted to improve upon the architectures presented in this paper. PororoGAN \cite{zeng2019pororogan} aims to improve the semantic relevance and overall quality of the images via a variety of textual alignment modules and a patch-based image discriminator. \citet{LI2020102956} also improve upon the StoryGAN architecture by upgrading the story encoder, GRU network, and discriminators and adding Weighted Activation Degree \cite{WEN2019251}. \citet{song2020CPCSV} is a more recent work which makes improvements to the StoryGAN architecture; the primary contribution is adding a figure-ground generator and discriminator, which segments the figures and the background of the image. Our model improvements of MART, dual learning, and copy-transform build upon more recent techniques and we support them with a detailed series of ablations.

\paragraph{Text-to-Image and Text-to-Video Generation.}
While story visualization is an underexplored task, there has been plenty of prior work in text-to-image synthesis. Most papers in this area can be traced back to StackGAN \cite{han2017stackgan}. Subsequent work then made various modifications to this architecture, adding attention mechanisms, memory networks, and more \cite{xu2018attngan,dmgan,objgan19,yi2017dualgan,gao2019perceptual}. \citet{huang2018turbo} and \citet{qiao2019mirrorgan} are direct precursors of our work. Both of these works frame the generated output as an image captioning task which attempts to reproduce the original text. Our proposed dual learning approach is an expansion of this module, where we use a state-of-the-art video captioning model based upon the MART \cite{lei2020mart} architecture to provide an additional learning signal to the model and increase the semantic consistency across images.

In the domain of text-to-video synthesis, \citet{li2018video}, \citet{pan2017to}, \citet{gupta2018imagine} and \citet{ijcai2019-276} generate videos from single sentences. In contrast to videos, story visualization does not have the requirement that the frames flow continuously together. Therefore, it allows for more interesting interactions and story-level dynamics to be captured that would only be present in longer videos.

\paragraph{Interactive Image Editing.}
Another task related to story visualization is interactive image editing. In this setting, rather than going from purely text to image, the model is given an input image along with textual instructions/directions, and must produce an output image that modifies the input image according to the text. This can take the form of high level semantic changes to the image, such as color and shape, as in \citet{describe_what_to_change}, \citet{nam2018text}, and \citet{chen2018language}, or this might take the form of Photoshop-style edits, as in \citet{pixeltone}, \citet{shi_image_editing}, and \citet{MANUVINAKURIKE18.481}. Alternatively, \citet{seq_gan_iie}, \citet{manuvinakurike2018conversational}, and \citet{el2019tell} are slightly closer to our task due to their sequential nature, where an image is modified repeatedly according to the textual feedback provided via a dialogue. However, unlike story visualization, these tasks do not have a narrative component. Furthermore, they involve repeatedly editing a single object at each timestep instead of generating diverse scenes with dynamic characters.

\begin{figure*}[t]
\centering
\includegraphics[width=0.95\textwidth]{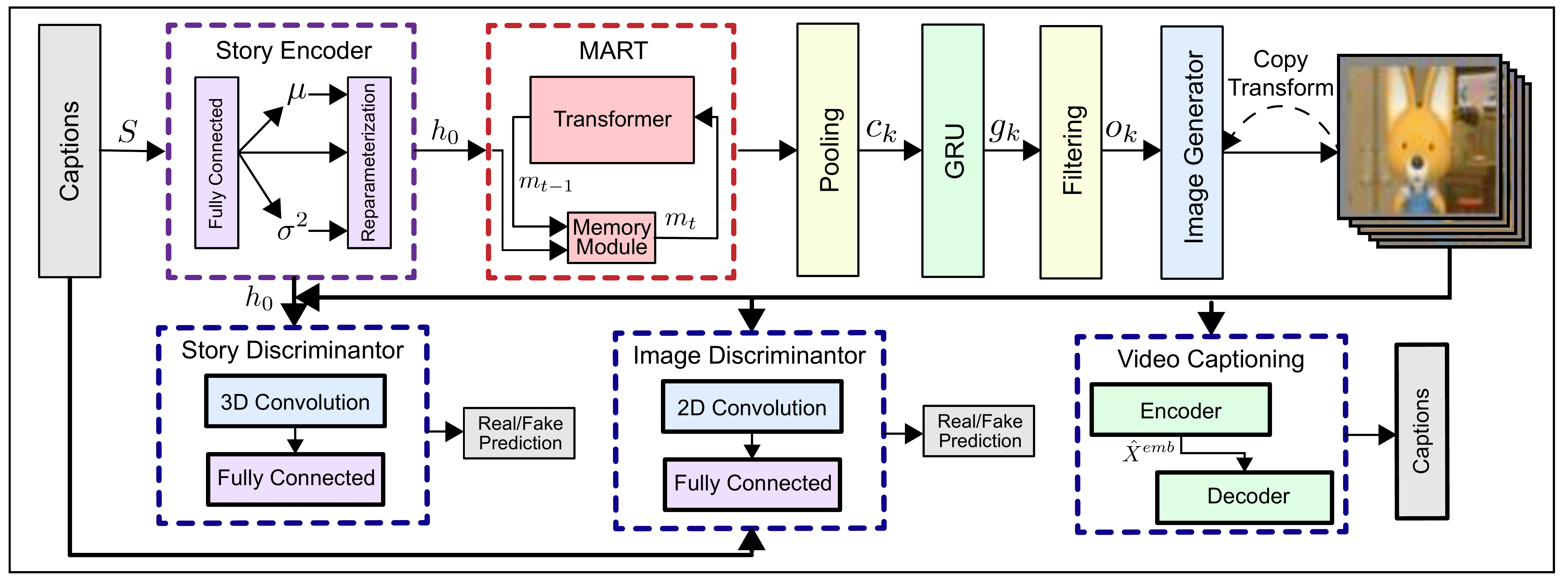}
\vspace{-7pt}
\caption{Illustration of \modelname{} architecture. The story encoder is used to initialize the memory module in the MART context encoder, which encodes the captions for the image generator. The copy-transform mechanism copies features from the images generated in previous timesteps to the image in current timestep. The generated images are passed to the story and image discriminators, and the dual learning video captioning model. \vspace{-3pt}
\label{fig:model_architecture}}
\vspace{-10pt}
\end{figure*}

\section{Methods}
\label{sec:methods}
\subsection{Background}
Formally, the task consists of a sequence of sentences $S=[s_1, s_2, ..., s_T]$ and a sequence of images $X=[x_1, x_2, ..., x_T]$, where the sentence $s_k$ describes the contents of the image $x_k$. The model receives $S$ as input and produces a sequence of images $\hat{X} = [\hat{x}_1, \hat{x}_2, ..., \hat{x}_T]$, attempting to accurately reproduce $X$. As detailed in \citet{li2019storygan}, there are two aspects of this task.
The first is local consistency, which is concerned with the quality of individual pairs in the sequence; an example is locally consistent if image $\hat{x}_k$ accurately represents the contents of sentence $s_k$. The second aspect is global consistency, which is concerned with the quality of the entire sequence. Namely, whether the sequence of images $\hat{X}$ accurately captures the content of the sequence of sentences $S$.

The general approach to this task as followed by StoryGAN \cite{li2019storygan} is as follows:
The story encoder creates the initial representation $h_0$ of the story $S$. This is then passed to the context encoder, which is a recurrent model that takes a sentence $s_k$ as input and forms a representation $o_k$. Each of these representations $o_k$ are then fed to the image generator, which outputs an image $\hat{x}_k$. The generated images are passed to two discriminators, the image discriminator and story discriminator, which each evaluate the generated images $\hat{x}_k$ in different ways and produce a learning signal that can be used to adjust the parameters of the network.

\subsection{\modelname{}}
The framework of our model is based on the StoryGAN architecture. We improve upon the context encoder and expand the network with dual learning and copy-transform mechanisms. The image and story discriminators, and the story encoder from the original model are retained in \modelname; each contributes to a separate loss term i.e. $\mathcal{L}_{img}$, $\mathcal{L}_{story}$ and $\mathcal{L}_{KL}$ respectively. See Appendix for details on the loss terms. An overview of our model architecture can be seen in Figure \ref{fig:model_architecture}.

\paragraph{MART Context Encoder.}
 One of the primary challenges of story visualization is maintaining consistent background imagery and character appearances throughout the story. This is addressed with a recurrent context encoder which has access to the global narrative while encoding the caption in each time-step. We use the Memory Augmented Recurrent Transformer (MART) \cite{lei2020mart}, where the memory is initialized with the conditioning vector $h_0$ from the story encoder. It takes word embeddings $W_k = [w_{k1}, w_{k2},....w_{kL}]$ where $w_{ij}\in\mathcal{R}^{1\times d_w} $, corresponding to the frame caption at each timestep and produces contextualized embeddings which are then pooled to a single weighted representation $c_k$ using attention.  This allows the context encoder to capture sophisticated interactions among the words which the image generator can then capitalize on:
\begin{gather*}
[m_{k1},....m_{kL}], h_{k} = \text{MART}([w_{k1},....w_{kL}], h_{k-1})\\
c_k = \sum_{i=1}^{L}\alpha_{ki}m_{ki}; \> \alpha_{ki} = \frac{exp(m_{ki}^{T}u)}{\sum exp(m_{ki}^{T}u)}
\end{gather*}
where $u$ is a query vector learned during training. The Transformer encoder is followed by a layer of GRU cells that take the contextualized embedding as input along with isometric Gaussian noise, $\epsilon_k$, and produce an output vector $g_k$. The outputs $c_k$ and $g_k$ are concatenated and transformed into filters, and subjected to convolution with a projection of the sentence embedding $s_k$, resulting in output vector $o_{k}$. See Appendix for more details.

\paragraph{Image Generator.}
The image generator follows prior text-to-image generation approaches \cite{qiao2019mirrorgan,xu2018attngan,han2017stackgan} and uses a two-stage approach. The first stage uses outputs $o_k$; the resulting image is fed through a second stage, which aligns the contextualized word encodings $m_{k}$ from MART with image sub-regions generated in first-stage and reuses weighted encodings for image refinement.

\paragraph{Dual Learning via Video Redescription.}
Dual learning provides the model with an additional learning signal by taking advantage of the duality of certain tasks, i.e., if X can be used to produce Y, then Y can be used to produce X. Here, our primary task is story visualization, and we consider the secondary task of video captioning. We refer to this process as video redescription.
To execute the idea of learning via video redescription, we employ a video captioning network which takes the sequence of generated images and produces a corresponding sequence of captions. The video captioning network is based on a recurrent encoder-decoder framework ($V_{enc}(.)$, $V_{dec}(.)$) and is trained using a cross-entropy loss on the predicted probability distribution ($p$) over its vocabulary. Specifically, $\mathcal{L}_{dual} = \sum_{k=1}^{T}\sum_{i=1}^{L}\text{log}p_{ki}(w_{ki})$. The hidden state in the recurrent model helps the captioning network to identify narrative elements in the sequence of images and penalize the generative model for a lack of consistency in addition to semantic misalignment. We pretrain the video captioning network using ground truth data and freeze its parameters while training the generative model. We also include a multiplier, $\lambda_{dual}$, which allows us to scale the loss.
The implementation of the encoder-decoder framework can vary. For our primary model, we adapt the MART video captioning network \cite{lei2020mart} to accept a 2D matrix of features at each time step where each column corresponds to an image sub-region (see Sec.~\ref{sec:evaluation}).

\paragraph{Sequentially-Consistent Story Visualization.}

While certain components, such as character positions, will change from frame to frame, there are other components like background and appearances which usually carry over to adjacent frames. To take advantage of this continuity, we augment the model with a copy-transform mechanism. This mechanism can take into consideration the generated image from previous timesteps, and reuse aspects of those prior images during the current timestep. The copy-transform module $F^{copy}(.)$ performs attention-based semantic alignment \cite{xu2018attngan} between word features $m_{k}\in\mathcal{R}^{D_{w}\times L}$ in the current timestep and image features $i_{k-1}\in\mathcal{R}^{D_{i}\times N}$ from previous step. Each column of $i_{k-1}$ is a feature vector of a sub-region of the image. The word features are first projected into the same semantic space as image features i.e. $m'_{k} = Um_{k}$, where $U\in \mathcal{R}^{D'\times D}$. For the $j^{th}$ image sub-region, the word-context vector is calculated as:
\begin{gather*}
    c_{jk} = \sum_{i=0}^{L}\beta_{ji}m'_{ik};\>\> \beta_{jik}=\frac{\text{exp}(h_{j}^{T}m'_{ik})}{\sum_{i=0}^{L}\text{exp}(h_{j}^{T}m'_{ik})}
\end{gather*}

$\beta_{jik}$ indicates the weight assigned by the model to the $i^{th}$ word when generating the $j^{th}$ sub-region of the image. The weighted word-context matrix is then concatenated with the generative image features from the current timestep and sent for upsampling to the image generator.

\paragraph{Objective.} Bringing it all together, the final objective function of the generative model is:
\begin{equation*}
    \min_{\theta_{G}} \max_{\theta_{I},\theta_{S}} \> \mathcal{L}_{KL} + \mathcal{L}_{img} + \mathcal{L}_{story} + \lambda_{dual}\mathcal{L}_{dual}
\end{equation*}

where $\theta_{G}$, $\theta_{I}$ and $\theta_{S}$ denote the parameters of the entire generator, and image and story discriminator respectively. See Appendix for more details.

\section{Experiments}
\label{sec:experiments}

\paragraph{Dataset.}
We utilize the Pororo-SV dataset from the original StoryGAN paper which has been adapted from a video QA dataset based on an animated series \cite{li2019storygan}\footnote{We opt to not use the CLEVR-SV dataset as we believe that this dataset lacks a narrative structure and is not suitable for story visualization.}. Each sample in Pororo-SV contains 5 consecutive pairs of frames and captions. The original splits of Pororo-SV from \citet{li2019storygan} contain only training and test splits with nearly 80\% overlap in individual frames. For a more challenging evaluation, we use the test split proposed in \cite{li2019storygan} as the validation split (2,334 samples) and carve out an "unseen" test split from the training examples. The resulting dataset contains 10191, 2334 and 2208 samples in training, validation and test splits respectively. In this version, there is 58\% frame overlap between the validation and train splits and 517 samples in the validation split contain at least one frame which is not present in the training set. Conversely, the test split has zero overlap with the training split.

\paragraph{Experimental Settings.}
Our model is developed using PyTorch, building off of the original StoryGAN codebase. All models are trained on the proposed training split and evaluated on the validation and test sets. We select the best checkpoints and tune hyperparameters by using the character classification F-Score on validation set (see Appendix).

\section{Evaluation of Visual Story Generation}
\label{sec:evaluation}
As with any task, evaluation is a critical component of story visualization; however, due to the complexity of the task and its generative nature, evaluation is nontrivial. For instance, characters are the focal point of any narrative and similarly should be the focus of a model when producing images for the story. Hence, \citet{li2019storygan} measure the character classification accuracy within frames of generated visual stories in order to compare models. However, it is also important that the characters and background are consistent in appearance, and together form a cohesive story rather than an independent set of frames. Inspired by insights such as this, we explore an additional set of evaluation metrics that capture diverse aspects of a model's performance on visual story generation.

\begin{figure}[t]
\centering
\includegraphics[width=0.47\textwidth]{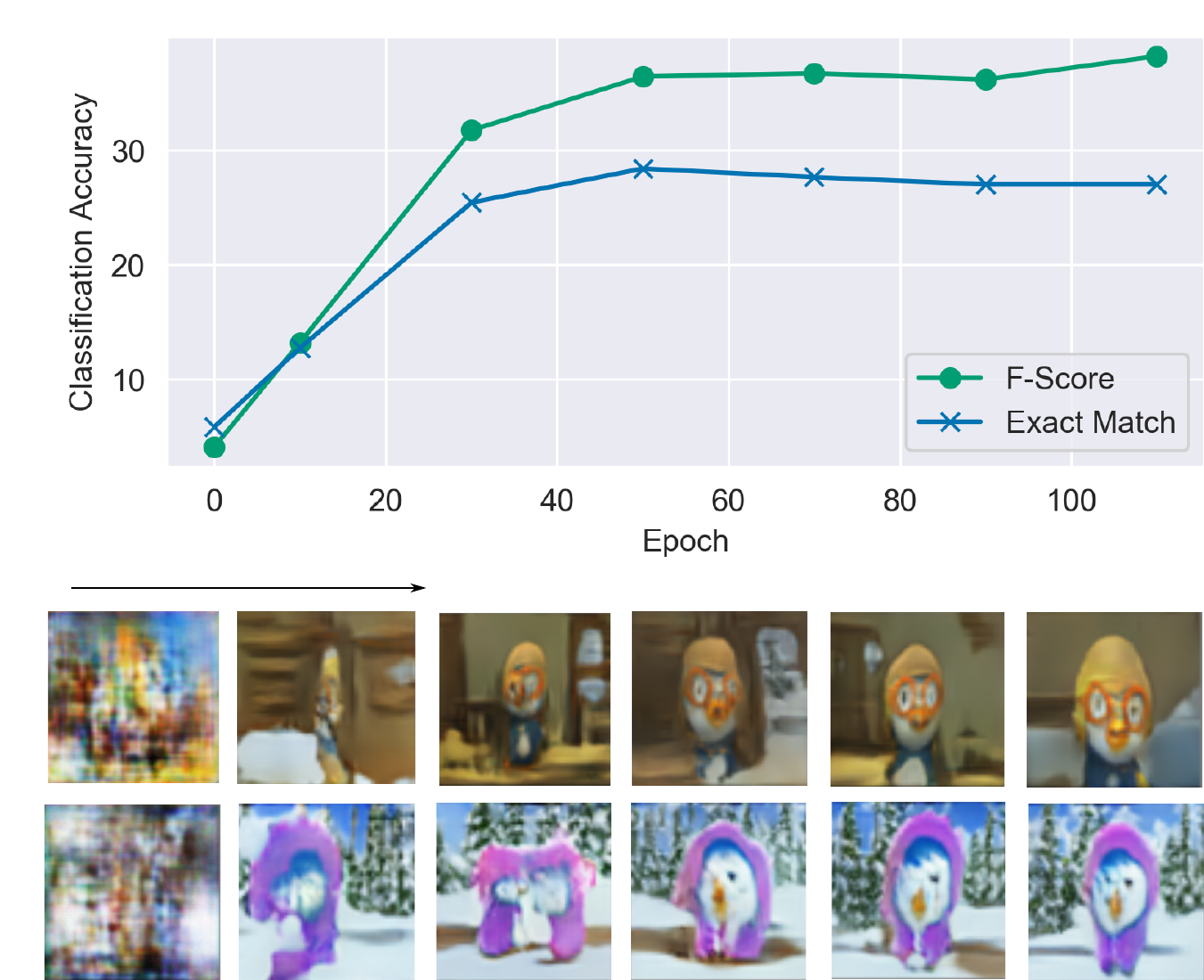}
\caption{Progression of character classification scores (top) and generated images (bottom) with training.
\label{fig:epochs}}
\vspace{-10pt}
\end{figure}

\textbf{Character Classification.} We finetune the pretrained Inception-v3 \cite{szegedy2016rethinking} with a multi-label classification loss to identify characters in the generated image. Most earlier work in story visualization report the image-level exact-match (EM) character classification accuracy. However, we contend that the exact match accuracy is not sufficient to gauge the performance of generative models, and the micro-averaged F-score of character classification should also be reported. For example, if Model A generates one of two characters in a frame with better quality than Model B (which generates none), it results in the same EM accuracy as Model B but an improvement in the recall/F-Score of the model, making the latter more reliable as a metric for quality. Our conclusion is based on the observation of consistent improvement in character classification scores with increasing training epochs and manual evaluation of image quality (see Fig.~\ref{fig:epochs}).

\textbf{Video Captioning Accuracy.} In order to measure global semantic alignment between captions and generated visualizations, we propose to use video captioning models which have been pretrained on ground truth data to identify narrative elements in a sequence of frames. We use the Memory-Augmented Recurrent Model proposed in \citet{lei2020mart} and add a CNN encoder \cite{sharma2018conceptual} on top of the Transformer encoder to extract image embeddings. The final convolutional layer (\texttt{Mixed\_7c}) in finetuned Inception-v3 is used to extract a local feature matrix $f \in \mathcal{R}^{64\times 2048}$ (reshaped from $2048\times8\times8$) for each image in the story. We then use this trained video captioning model to caption the generated frames. The generated captions are compared to the ground truth captions via BLEU evaluation\footnote{We use the \texttt{nlg-eval} package \cite{sharma2017nlgeval} for BLEU evaluation.}, and this functions as our proposed metric for measuring global semantic-alignment between the captions and generated story. This pretrained model is also used as the video captioning dual network during training of \modelname.

\textbf{Discriminative Evaluation.}
Generative metrics such as BLEU are known to be noisy and unreliable. Hence, we also develop a discriminative evaluation setup. In order to compute similarity between the generated image and ground truth, we compare the feature representations from both images in this discriminative setup. The training dataset for story visualization may contain one or more frames with the exact set of characters that are referenced in captions in the evaluation data. When we are checking for the presence of these characters in a generated image, we do not want to reward the model for copying the exact same frame from the training set instead of generating a frame suited to the input caption. In order to evaluate this consistency, we propose discriminative evaluation of the story visualization model. Using the character annotations for the final frame of each sample in the test splits, we extract a set of 4 negative frames which are taken from elsewhere in the video but contain those specific characters (see Fig.~\ref{fig:pororo_disc} in Appendix). The human evaluation accuracy on this dataset is 89\% ($\kappa$=0.86) and is used as an upper bound when interpreting model accuracy performance. The cosine similarity between Inception-v3 features of final generated frame and candidate frames is computed and the frame with the greatest similarity is selected as predicted frame. We report Top-1/2 accuracies.

\renewcommand{\arraystretch}{1.3}%
\begin{table*}
\small
\centering
\begin{tabular}{|c|c|c|c|c|c|c|}
\hline
\textbf{Model} & \textbf{Char. F1} & \textbf{BLEU2/3} & \textbf{R-Precision} & \textbf{Frame Acc.} & \textbf{Top-1 Acc.} & \textbf{Top-2 Acc.}\\ 
\hline
StoryGAN \cite{li2019storygan} & 41.11 & 3.86 / 1.72 & 3.40 $\pm$ 0.01 & 21.90 & 22.42 & 45.40 \\
\hline 
StoryGAN + Transformer & 42.45 & 3.92 / 1.73 & 4.03 $\pm$ 0.17 & 22.14 & 23.79 & 47.15 \\
\hline 
CP-CSV \cite{song2020CPCSV} & 43.79 & 3.96 / 1.73 & 3.97 $\pm$ 0.21 & 22.08 & 24.29 & 46.39 \\
\hline
  \modelname & \textbf{48.27} & \textbf{4.51 / 1.92} & \textbf{6.10 $\pm$ 0.07} & \textbf{22.71} & \textbf{25.62} & \textbf{47.39} \\
\hline
\end{tabular}
\vspace{-5pt}
\caption{\label{tab:pororo_test_1} Results on validation split of Pororo-SV Dataset.}
\vspace{-5pt}
\end{table*}

\renewcommand{\arraystretch}{1.3}%
\begin{table*}
\small
\centering
\begin{tabular}{|c|c|c|c|c|c|c|}
\hline
\textbf{Model}  & \textbf{Char. F1} & \textbf{BLEU2/3} & \textbf{R-Precision} & \textbf{Frame Acc.} & \textbf{Top-1 Acc.} & \textbf{Top-2 Acc.}\\ 
\hline
StoryGAN \cite{li2019storygan} & 18.59 & 3.24 / 1.22 & 1.51 $\pm$ 0.15 & 9.34 & 23.14 & 42.27 \\
\hline
StoryGAN + Transformer & 19.29 & 3.29 / 1.23 & 1.49 $\pm$ 0.07 & 9.58 & 23.31 & 42.29 \\
\hline
CP-CSV \cite{song2020CPCSV}  & 21.78 & 3.25 / 1.22 & 1.76 $\pm$ 0.04 & 10.03 & 22.23 & 41.86 \\
\hline
  \modelname & \textbf{38.01} & \textbf{3.68 / 1.34} & \textbf{3.56 $\pm$ 0.04} & \textbf{13.97} & \textbf{23.72} & \textbf{42.48} \\
\hline
\end{tabular}
\vspace{-5pt}
\caption{\label{tab:pororo_test_2} Results on test split of Pororo-SV Dataset.}
\vspace{-10pt}
\end{table*}

\textbf{R-Precision.} Several prior works on text-to-image generation report the retrieval-based metric R-Precision \cite{xu2018attngan} for quantifying the semantic alignment between the input text and generated image. If there are $R$ relevant documents for a query, the top $R$ ranked retrieval results of a system are examined; if $r$ are relevant, the R-precision is $r/R$. In our task\footnote{The R-precision score is obtained from 10 runs with 99 randomly picked mismatched story candidates in each run.}, $R=1$. The encodings from a pretrained Deep Attention-based Multimodal Similarity Model (DAMSM) are used to compute cosine similarity and rank results. Since this model only evaluates a single text-image pair for similarity, it is not suitable for evaluating story visualization. Therefore, we train a new version of DAMSM to extract global representations for the story and sequence of images, referred to as Hierarchical DAMSM (H-DAMSM) (see Appendix).

The models used in the aforementioned evaluation metrics are trained independently of \modelname{} on the proposed Pororo-SV splits and the pretrained weights are used for evaluation. See Appendix for upper bounds.

\begin{figure}[t]
\centering
\includegraphics[width=0.47\textwidth]{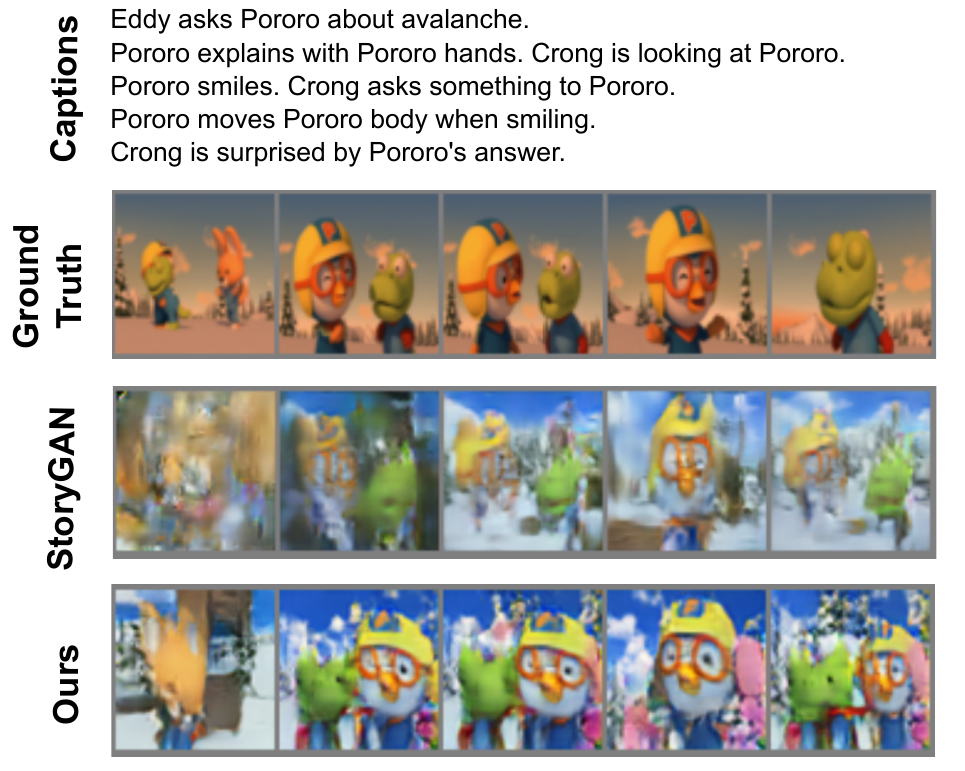}
\caption{Sample results from StoryGAN and \modelname{} on unseen test split.
\label{fig:test_samples}}
\vspace{-7pt}
\end{figure}

\section{Results}
\label{sec:results}

\subsection{Main Quantitative Results}
The results for the Pororo-SV validation set can be seen in Table \ref{tab:pororo_test_1}. The first row contains the results using the original StoryGAN model \cite{li2019storygan}\footnote{We use a reduced training dataset as compared to the original StoryGAN paper (see Sec~\ref{sec:experiments}). However, we evaluate our StoryGAN code base on their exact splits and get 26.1\% exact-match accuracy, which is approximately equivalent to the 27\% reported in the original paper where they demonstrate that StoryGAN outperforms previous baselines such as ImageGAN, SVC, and SVFN.}. The second row functions as another baseline, where we replace the GRU-based context encoder in StoryGAN with a Bidirectional Transformer \cite{devlin2019bert}. The conditioning augmentation vector is not used to initialize the context encoder in this model since a non-recurrent Transformer lacks a hidden state. We see 1-2\% improvements in character classification and retrieval with this model over StoryGAN. The third row contains results from the more recent CP-CSV model \cite{song2020CPCSV} which uses figure-ground segmentation as an auxiliary task for preserving character features. Consequently, it results in 2.68\% improvement in character classification over StoryGAN and smaller improvements for other metrics. The final row contains results with \modelname{}, which significantly outperforms previous models (including CP-CSV) across all metrics. The character classification F-Score improves by 7.16\% suggesting that the characters generated in our images are of higher visual quality. Similarly, we see consistent improvements in BLEU as well as R-Precision with our model. As demonstrated in Sec~\ref{sec:ablations}, the improvement in BLEU can be attributed to the addition of dual learning, which directly optimizes the dual task of video captioning. The R-Precision indicates that our model learns better global semantic alignment between the captions and images. Lastly, the Top-1/2 accuracy scores show that our model is learning to generate diverse images, rather than copying scenes that feature the same characters from the training data.

\modelname{} performs dramatically better than other models on the unseen test split (see Table~\ref{tab:pororo_test_2}). As can be seen in Fig~\ref{fig:test_samples}, StoryGAN performs rather poorly on unseen samples compared to \modelname{}. While the former produces images that are blurry and character shapes that are faint, the latter generates frames with sharp character features. This is reflected in the wide improvement margins on character classification in Table~\ref{tab:pororo_test_2}. Similar improvements are also observed for BLEU and R-Precision metrics, indicating that our model generates images which are more relevant to the input caption. When generating stories for the Pororo-SV test split, models tend to copy background elements from the samples seen in the training set, since the captions lack sufficient information about the setting. Hence, we observe little improvement over random chance in the discriminative accuracy scores for different models on the test split. For instance, instead of generating the tinted background in the ground truth in Fig.~\ref{fig:test_samples}, the models produce a clear blue sky which is closer to samples seen in the training set. However, discriminative evaluation will be valuable for future work in this domain when inputs contain detailed information about the visual elements.

We also provide per character results for the Character F-Score. With \modelname{}, we see up to 20\% improvement for less frequent characters (see Table~\ref{tab:char_scores}).

\subsection{Human Evaluation}

\begin{table}
\small
\centering
\begin{tabular}{|c|c|c|c|c|}
\hline
 & \multicolumn{2}{c|}{Win \%} & \multicolumn{2}{c|}{Mean Rating} \\ 
\hline
Attribute & Ours & StGAN & Ours & StGAN \\
\hline
Visual Quality & \textbf{82\%} & 3\% & \textbf{2.06} & 1.22\\
Consistency & \textbf{78\%} & 3\% & \textbf{2.94} & 1.78 \\
Relevance & \textbf{26\%} & 2\% & \textbf{1.28} &  1.04 \\

\hline
\end{tabular}
\caption{\label{tab:human_eval} Human evaluation on Likert Scale 1-5. Win\% = \% times stories from one model was preferred over the other (StGAN = StoryGAN). Tie\% = \% samples remaining after considering Win\% of both models.}
\vspace{-15pt}
\end{table}

We conduct human evaluation on the generated images from \modelname{} and StoryGAN, using the three evaluation criteria listed in \citet{li2019storygan}: visual quality, consistency, and relevance. Two annotators are presented with a caption and the generated sequence of images from both models, and asked to rate each sequence on a scale of 1-5. Results are presented in Table \ref{tab:human_eval}. With respect to pairwise evaluation, predictions from our model are nearly always preferred over those from StoryGAN (see Win\% columns). Similarly, we see large improvements in mean rating of stories generated by \modelname{}. However, we also see higher Tie\% and low mean rating for the attribute Relevance, suggesting that much work remains to be done to improve understanding of captions.

\paragraph{Correlation Experiments. }We also examine the correlation between our proposed metrics and human evaluation of generated images. We compute the Pearson's correlation coefficient between human ratings of 50 samples on three different attributes using the 1-5 Likert scale and their corresponding automated metric evaluation scores. Significant correlation ($\rho=0.586$) was observed between our proposed Character F-Score metric and Visual Quality, lending strength to its use as an automated metric for story visualization.

\renewcommand{\arraystretch}{1.3}%
\begin{table*}
\small
\centering
\begin{tabular}{|c|c|c|c|c|c|c|}
\hline
\textbf{Model}  & \textbf{Char. F1} & \textbf{BLEU2/3} & \textbf{R-Precision} & \textbf{Frame Acc.} & \textbf{Top-1 Acc.} & \textbf{Top-2 Acc.}\\ 
\hline
StoryGAN \cite{li2019storygan} & 41.11 & 3.86 / 1.72 & 3.40 $\pm$ 0.01 & 21.90 & 22.42 & 45.40 \\
\hline
StoryGAN + Transformer  & 42.45 & 3.92 / 1.73 & 4.03 $\pm$ 0.17 & 22.14 & 23.79 & 46.15 \\
StoryGAN + MART  & 47.03 & 4.15 / 1.81 & 5.11 $\pm$ 0.12 & 22.25 & 24.48 & 46.42 \\
  + Story Captioning  & 47.23 & 4.78 / 1.87 & 6.32 $\pm$ 0.08 & 22.30 & 24.53 & 47.41 \\
  + Copy Transform  & 48.27 & 4.51 / 1.92 & 6.10 $\pm$ 0.07 & 22.71 & 25.62 & 47.39 \\
\hline
\end{tabular}
\caption{\label{tab:pororo_ablation} Ablation results on validation split of Pororo-SV dataset.}
\vspace{-10pt}
\end{table*}

\section{Discussion}
\subsection{Ablations}
\label{sec:ablations}
Table \ref{tab:pororo_ablation} contains plus-one ablations for \modelname. The first row is the StoryGAN baseline and the second row is the StoryGAN + Transformer model, as discussed in Section \ref{sec:results}. We then iteratively add each of our contributions and observe the change in metrics\footnote{Statistical significance is computed with 100K samples using bootstrap \cite{noreen1989computer, tibshirani1993introduction}. All our improvements in \modelname{}  are statistically significant, except for discriminative evaluation, and frame accuracy scores for the dual learning module.}. First, we upgrade the Transformer encoder to MART, which brings about the largest improvements across all metrics. The use of word embeddings with access to the global conditioning vector and attention-based semantic alignment proves important to the task of story generation. Next, we use the MART context encoder with our proposed dual learning and copy-transform improvements. With the addition of video captioning as a learning signal, we see 0.20\% (p=0.071) improvement in character F-score and 1.12\% improvement in R-Precision (p=0.032) over MART. The highest improvements are observed for BLEU score, since the model is optimized on video captioning. Next, we evaluate the addition of the copy-transform mechanism where features from generated images in previous timesteps are copied to the image in the current timestep. We observe 1.04\% improvements for character classification and a slight drop in performance on video captioning. Similarly, there is 1.14\% improvement in Top-1 accuracy for the discriminative dataset.

As discussed in Section \ref{sec:methods}, we explore a variety of implementations for the dual learning component of our model. While MART-based video captioning works the best, we provide a discussion of other approaches in the Appendix.

\begin{figure}[t]
\centering
\includegraphics[width=0.46\textwidth]{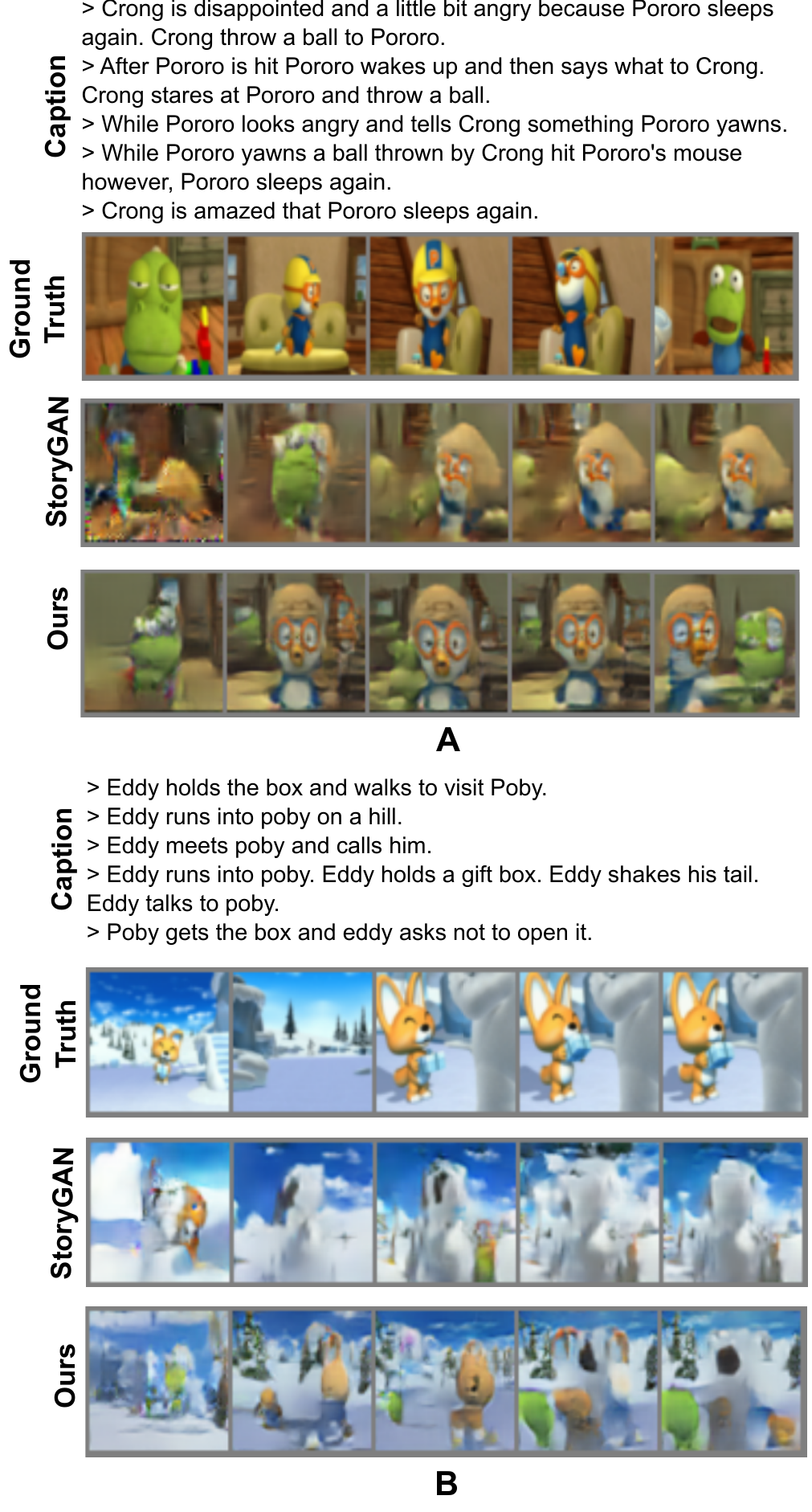}
\vspace{-10pt}
\caption{Comparative examples of generated images.
\label{fig:examples}}
\vspace{-10pt}
\end{figure}

\subsection{Qualitative Examples}
Figure \ref{fig:examples} contains two generated examples from the Pororo-SV dataset. The top row in each example contains the ground truth images, the middle row the images generated by StoryGAN, and the final row the images generated by our model. In example A, we demonstrate the superior visual quality and consistency of the frames generated by \modelname{}, as compared to StoryGAN. The MART encoder allows our model to comprehend long captions as well as attend to each word while generating images. The retention of native character features throughout the story during regeneration can be attributed to the copy-transform mechanism in our model. In contrast, we see that both models fail at generating defined characters in example B. This may be due to the fact that \textit{Poby} is an infrequent character in the dataset and hence, both models fail to learn its features.

\subsection{Linguistic Analysis}
We perform a visual analysis of the captions and predictions from \modelname{} and observe two major recurring themes. First, the frequency of characters in the training data is a significant deciding factor for generated image quality. We looked at the samples that contained at least \textit{Pororo} (most frequent character) and found that generated stories are better when there is only a single character in the frame’s narrative as compared to multiple characters. This points to the inability of current story visualization models to align captions with multiple subjects/objects to the corresponding images. Second, generated images are poor for scenes containing infrequently occurring objects such as book/plate/boat/plane etc. in the caption. This behavior is expected since the model is unaware of real-world objects that do not already appear in the training set with sufficient frequency. Moreover, since the Pororo-SV dataset has been adapted from the annotations of a video QA dataset, the captions often contain information that can only span over multiple frames (\say{Pororo wakes up and then says what to Crong. Pororo stares at Pororo and throws a ball}), or cannot be visualized through images (\say{Poby gets the box and Eddy asks not to open it.}). Hence, our results with metrics like BLEU and R-Precision, which are supposed to capture the relevance between images and captions, stay relatively low (see Tables~\ref{tab:pororo_test_1} and ~\ref{tab:pororo_test_2}).

\section{Conclusion}
In this paper, we investigate the underexplored task of story visualization. We improve upon prior modeling approaches and demonstrate the effectiveness of these new approaches by performing a robust set of ablation experiments. We also present a detailed set of novel evaluation methods, which we validate by demonstrating improvements across various baselines. Evaluation for story visualization is a challenging open research question in itself, and we hope that these methods will encourage more work in this domain.

\section{Ethics/Broader Impacts}
\label{sec:ethics}
From an ethics standpoint, we provide a brief overview of the data that the model is trained on in Section \ref{sec:experiments} and a more detailed discussion in the Appendix. We provide some analyses of the data and refer the reader to the original StoryGAN paper, where the dataset was created, for further details. All of the language data consists of simple English sentences. Our experimental results are specific to the story visualization task. Pororo-SV is the most challenging story visualization task available; therefore, our results would likely generalize to other story visualization datasets. While story visualization is an exciting task with many potential future applications, the generated images still contain many obvious visual artifacts and therefore models trained on this task are still far from being deployed in any real world settings.

Story visualization minimizes many of the ethical issues associated with image and video generation. DeepFakes, which are algorithmically generated fake images, have become increasingly problematic \cite{nguyen2019deep}. Oftentimes, these images are indistinguishable from real images, raising privacy concerns and providing a source of misinformation. The images that we generate here are not subject to this same issue, due to the fact that they are Cartoons, and are therefore unable to be confused with real images. The focus of the task is not on the realism of the images, but rather on the multimodal narrative. Therefore, cartoons are actually better suited for the task as real-world images only add additional visual complexity that is not relevant to the narrative.

\section*{Acknowledgments}
We thank Peter Hase, Jaemin Cho, Hyounghun Kim, and the reviewers for their useful feedback. This work was supported by DARPA MCS
Grant \#N66001-19-2-4031, DARPA KAIROS Grant \#FA8750-19-2-1004, ARO-YIP Award W911NF-18-1-0336, and a Google Focused Research Award. The views are those of the authors and not of the funding agency.

\bibliography{naacl2021}
\bibliographystyle{acl_natbib}

\appendix

\section*{Appendices}
\section{Methods}

StoryGAN only uses pretrained sentence embeddings as a representation for the caption, while \modelname{} uses a combination of sentence and word embeddings. The context encoder is responsible for encoding the captions and transforming them into image embeddings.

\paragraph{Story Encoder.}
The story encoder $E(.)$ encodes the entire story, $S$ into a single representation, $h_0$, which functions as the initial memory state of the MART model. The input $S$ is the concatenation of sentence embeddings $s_k\in\mathcal{R}^{1\times d_s}$ from all timesteps. The conditional augmentation technique \cite{han2017stackgan} is used to convert $S$ into a conditioning vector by using it to construct and sample a conditional Gaussian distribution i.e., $h_0 = \mu(S) + \sigma^2(S)^{1/2} \odot \epsilon_{S}$, where $\epsilon_{S} \sim \mathcal{N}(0, 1)$ and $\odot$ represents element-wise multiplication. This introduces a loss term which is the KL-Divergence between the learned distribution and Gaussian distribution i.e.,
\begin{equation*}
    \mathcal{L}_{KL} = KL(\mathcal{N}(\mu(S), \text{diag}(\sigma^2(S))) || \mathcal{N}(0, I))
\end{equation*}

\paragraph{Discriminators.}
There are two discriminators in the model, each aimed at capturing a different aspect of the task. The image discriminator focuses on local consistency and is provided with the generated image $\hat{x}_k$, the sentence $s_k$, and the context information vector from the story encoder $h_0$, and must attempt to distinguish between this and a real triplet, containing the same information except for the real image $x_k$ instead of the fake image ($\mathcal{L}_{img}$). Additionally, the image discriminator is also used to classify the characters in the frame, when labels are available. The story discriminator is instead concerned with the global consistency of the generated sequence. The generated image sequence $\hat{X}$ and story $S$ are provided to the discriminator, which must distinguish it from an equivalently encoded real pair ($\mathcal{L}_{story}$)

\paragraph{MART Context Encoder}
The MART encoder, as described in the main paper, is followed by a layer of GRU cells that take the contextualized embedding as input along with isometric Gaussian noise, $\epsilon_k$, and produce an output vector $g_k$. The outputs $c_k$ and $g_k$ are concatenated and transformed into filters, and subjected to convolution with a projection of the sentence embedding $s_k$ i.e.
\begin{gather*}
    g_k, \> q_{k} = \text{GRU}(s_k,\epsilon_k, q_{k-1})\\
    o_k = Filter([c_k;g_k]) \circ \text{tanh}(W_{I}s_k) 
\end{gather*}

where $q_k$ is the hidden state of the GRU cells. $Filter(.)$ transforms the concatenated vector $[c_k;g_k]$ into a multi-channel filter of size $C_{out}\times 1\times 1 \times len(W_{I}s_k)$, where $C_{out}$ is the number of output channels. The convolution operation can be interpreted as the sifting of information from the local context $s_t$ with the use of filters that have access to the global context.

\section{GAN Training}
The training procedure for our GAN architecture is similar to StoryGAN. The objective function for the generative model is:
\begin{equation*}
    \min_{\theta_{G}} \max_{\theta_{I},\theta_{S}} \> \mathcal{L}_{KL} + \mathcal{L}_{img} + \mathcal{L}_{story} + \lambda_{dual}\mathcal{L}_{dual}
\end{equation*}
where $\theta_{G}$ is the parameters of the generator, $\theta_{I}$ is the parameters for the image discriminator, and $\theta_{S}$ is the parameters for the story discriminator. Note that the video captioning dual learning component is pretrained and then frozen while the rest of the model is trained.

Each of the components in the model has a conditional loss, which is concerned with whether the input caption and generated image align. The adversarial loss function for the generator is then as follows:
\begin{align*}
    \mathcal{L}_{G_i} = 
    - \frac{1}{2} \mathrm{E}_{\hat{x}_{i} \sim p_{\hat{x}_i}}[log(D_{img}(\hat{x}_i, s))] \\
    - \frac{1}{2} \mathrm{E}_{\hat{X}_{i} \sim p_{\hat{X}_i}}[log(D_{story}(\hat{X}_i, S))]
\end{align*}
where $\hat{x}_i$ is the generated image sampled from the distribution $p_{\hat{x}_i}$ during the $i^{th}$ stage of generation. The first term is the conditional loss of the image discriminator, and the second term is the conditional loss for the story discriminator.

The adversarial losses for the discriminators are:
\begin{align*}
    \mathcal{L}_{D_{img}} = 
    -\frac{1}{2} \mathrm{E}_{x_{i} \sim p_{x_i}}[log(D_{img}(x_i, s))] \\
    -\frac{1}{2} \mathrm{E}_{\hat{x}_{i} \sim p_{\hat{x}_i}}[log(1 - D_{img}(\hat{x}_i, s))]
\end{align*}

\begin{align*}
    \mathcal{L}_{D_{story}} = 
    -\frac{1}{2} \mathrm{E}_{X_{i} \sim p_{X_i}}[log(D_{story}(X_i, S))]\\
    -\frac{1}{2} \mathrm{E}_{\hat{X}_{i} \sim p_{\hat{X}_i}}[log(1 - D_{story}(\hat{X}_i, S))]
\end{align*}
where $\hat{x}_i$ is the generated image sampled from the distribution $p_{\hat{x}_i}$, and $x_i$ is the real image sampled from the distribution $p_{x_i}$, during the $i^{th}$ stage of generation.

For additional algorithmic details we refer readers to \citet{li2019storygan}.

\section{Experimental Settings}
Our model is constructed using PyTorch, building off of the original StoryGAN codebase. All models are trained on the training set, tuned on the validation set, and evaluated on the test set. We report results for each of the latter. We select the best checkpoints and manual tune hyperparameters for each model by using the validation character classification F-Score. We use the ADAM optimizer with betas of 0.5 and 0.999. We train the model on a single Nvidia 2080TI GPU. Each epoch takes ~30 minutes, with the model being saved every 10 epochs. At 120 epochs of training, the total training time is nearly 60 hours for a batch size of 4. We did 1-5 runs for hyperparameter search using manual tuning. The number of trainable parameters in our proposed \modelname{} is 101,718,981.

\section{Hyperparameters}

Many of our hyperparameters are shared with the StoryGAN model. The image size that we use is 64-by-64, and the length of the story is 5 images/captions. The learning rate of the generator is 2e-4, while the learning rate of the discriminator is slightly lower at 1e-4. We train the model for 120 epochs and set the learning rate to decay every 20 epochs. For each training update of the discriminators, we perform two updates for the generator network, with different mini-batch sizes for image and story discriminators \cite{li2019storygan}. The image discriminator batch size is 20 and the story discriminator batch size is 4. We found in our experiments that all story visualization models are susceptible to mode collapse with small changes in the discriminator learning rate. Additionally, we attempted replacing the attention-based alignment module from \citet{xu2018attngan} with a cross-attention layer and observed mode collapse in later epochs for the first generated frame in the story. Usually, the mode collapse can be avoided with a higher batch size, but it does not lead to larger improvements. We also used an update ratio of 3:1 for generator vs. discriminator and did not find it useful.

The MART hyperparameters are as follows. The hidden size of the model is 192. The number of memory cells is 3. The number of hidden layers is 2. The dropout values across the model are 0.1. The layer normalization epsilon is 1e-12. The number of attention heads is 6. The word embedding size is 300, and the embedding is initialized using the 840B glove training checkpoint.

\section{Pororo-SV Dataset}
We utilize the Pororo-SV dataset from the original StoryGAN paper \cite{li2019storygan}. This dataset was originally a video QA dataset \cite{pororvoQA}, consisting of one second video clips paired with multiple descriptions. A sequence of these video clips forms a story, which then has QA pairs associated with it. There are 9 characters frequently featured in the dataset; a distribution of them can be seen in the supplementary. Annotations are available for the distribution of characters in each frame. It can be seen that each character is featured in at least 10\% of the frames, making it crucial for the model to be capable of generating each of them. To convert this to a story visualization task, \citet{li2019storygan} sample the one second videos, obtaining a single, representative frame. Five sequential frame-description pairs are then considered to make up a single story. We use the training and test splits outlined in \citet{li2019storygan} for comparable results. However, since this split is also used for tuning in both papers, we carve an equally-sized held-out split of unseen samples from the training set for fair evaluation of the models.

\begin{figure}[t]
\centering
\includegraphics[width=0.47\textwidth]{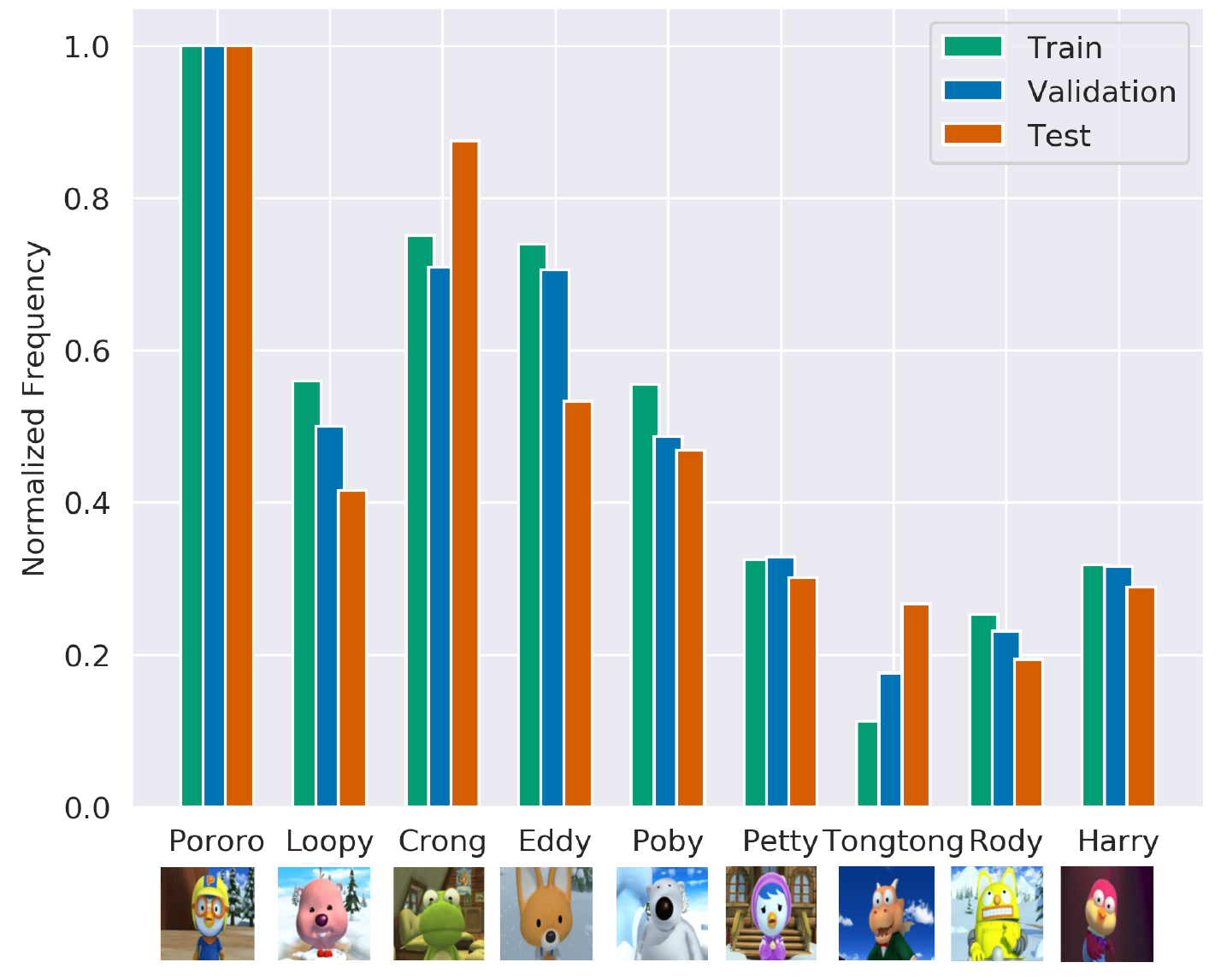}
\caption{Distribution of Pororo-SV characters in various data splits.
\label{fig:pororo_frequency}}
\end{figure}
\paragraph{Character Frequency. }The Pororo-SV dataset contains 9 characters that are frequently featured; a distribution of them can be seen in Figure \ref{fig:pororo_frequency}.

\section{Evaluation}
\paragraph{Video Captioning Accuracy.} Video Captioning models use a sequence of image embeddings from the sequence of frames in a video segment as input and perform decoding on the processed features to produce a caption consisting of single sentence or multiple sentences. However, they assume that there are multiple frames within a single video segment, unlike our story dataset where there is exactly one frame for each sentence in the story caption. Therefore, we adapt existing state-of-the-art video captioning models to perform decoding from a single frame for each sentence in the caption.

\textbf{R-Precision.} Several prior works on text-to-image generation report the retrieval-based metric R-Precision \cite{xu2018attngan} for quantifying the semantic alignment between the input text and generated image. R-Precision is computed using the similarity between encodings extracted from a pretrained Deep Attention-based Multimodal Similarity Model (DAMSM). Since this model only evaluates a single text-image pair for similarity, it is not suitable for evaluating story visualization. Therefore, we train a new version of DAMSM to encode all text-image pairs in each story and compute the global similarity for consecutive frames from a story and their respective captions in addition to sentence and word similarity. We introduce an additional bidirectional LSTM network for encoding frame captions into a story representation and average pool the image features for individual frames to extract a global visual embedding for the story. The cosine similarity between these two vectors is used to rank the retrieval-based search between the query visualization and candidate story narratives. This improved model, referred to as Hierarchical DAMSM (H-DAMSM), is trained using two additional story-level losses $\mathcal{L}_{st0}$ and $\mathcal{L}_{st1}$ with a smoothing coefficient of $\gamma$=15, and the pretrained model is used for evaluation. We refer the reader to \citet{xu2018attngan} for details on the DAMSM model.

\begin{figure}[t]
\centering
\includegraphics[width=0.47\textwidth]{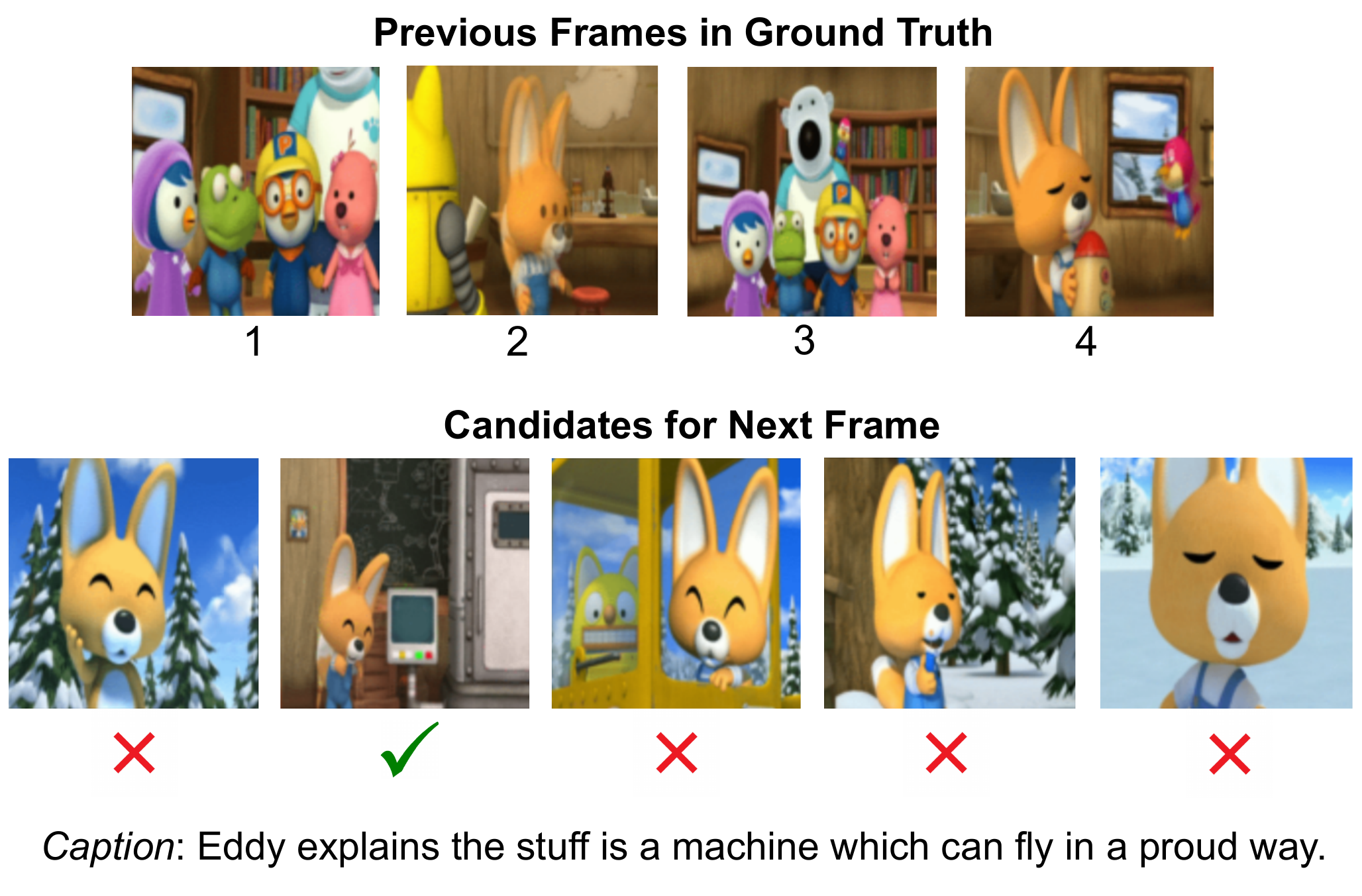}
\caption{Example of the Discriminative Dataset.
\label{fig:pororo_disc}}
\vspace{-10pt}
\end{figure}
\paragraph{Example of Discriminative Dataset.} Figure \ref{fig:pororo_disc} shows an example from our discriminative dataset that is used in the discriminative evaluation.

\section{Results}
\label{sec:sup_results}
\paragraph{Dual Learning.}
\renewcommand{\arraystretch}{1.3}%
\begin{table*}
\small
\centering
\begin{tabular}{|c|c|c|c|c|c|c|}
\hline
\textbf{ Dual Model} & \textbf{Char. F1} & \textbf{BLEU2/3} & \textbf{R-Precision}  & \textbf{Frame Acc.} & \textbf{Top-1 Acc.} & \textbf{Top-2 Acc.}\\ 
\hline
Image Captioning (CNN-LSTM) & 47.08 & 4.29 / 1.83 & 5.23 $\pm$ 0.06 & 22.29 & 24.47 & 46.48 \\
Video Captioning (CNN-LSTM) & 46.19 & 3.98 / 1.73 & 4.04 $\pm$ 0.29 & 22.12 & 23.93 & 46.22 \\
Image Captioning (Transformer) & 47.21 & 4.58 / 1.81 & 5.37 $\pm$ 0.11 & 22.47 & 24.47 & 46.51 \\
Video Captioning (Transformer) & 47.23 & 4.78 /1.87 & 6.32 $\pm$ 0.08 & 22.30 & 24.53 & 47.41 \\
\hline
\end{tabular}
\vspace{-5pt}
\caption{\label{tab:dual} Results from variations of Dual Learning on Pororo-SV dataset.}
\vspace{-10pt}
\end{table*}

The actual implementation of the encoder-decoder framework in our dual learning approach can vary. For our primary model, we adapt the MART architecture \cite{lei2020mart} to accept a 2D matrix of image features where each column corresponds to a sub-region, instead of a sequence of image features from adjacent frames in a video segment, for each time step (see details in the Evaluation section of the main paper). However, we compare this model with several variations of dual learning networks: (1) Transformer-based Image Captioning, (2) CNN-LSTM-based Video Captioning and (3) CNN-LSTM-based Image Captioning. The Transformer-based image captioning network is essentially a non-recurrent version of the MART-based video captioning. The CNN-LSTM based image captioning model is similar to \citet{qiao2019mirrorgan}. 
The generated image $\hat{x}_k$ is fed into a CNN, which produces a feature vector. The feature vector is then fed through an LSTM decoder, which produces the caption $\hat{s}_k$. The CNN-LSTM video-captioning model is an extension of this, using 3D convolutions to pool over all frames within a story. We pretrain these models on the Pororo-SV dataset and freeze the parameters before utlizing the weights to get the dual learning loss while training \modelname{}.

As seen in Table~\ref{tab:dual}, the image captioning approach using CNN-LSTM has a limited impact on performance. Next, we explore Transformer for implementing the captioning model and see larger improvements for character classification and BLEU scores. However, there is limited improvement in performance on R-Precision using image captioning as dual learning. We hypothesize that this is due to the image captioning model's inability to capture information across frames; essentially, this method of dual learning is only capable of considering local consistency and not global consistency. Therefore, we use a video captioning model, where all frames are considered simultaneously, allowing it to capture both local consistency and global consistency. The performance of CNN-LSTM based video captioning model on the captioning validation set was low. Hence, using this model for dual learning loss negatively affected performance of our story visualization model. The Transformer-based image-captioning model outperforms video-captioning with CNN-LSTM, suggesting that a sophisticated dual model is as important as global context for story visualization. Consequently, the MART-based video captioning model leverages additional global context and outperforms Transformer-based image captioning across all metrics.

\renewcommand{\arraystretch}{1.3}%
\begin{table}
\small
\centering
\begin{tabular}{|c|c|c|c|c|}
\hline
\textbf{Character} & \textbf{Support} & \textbf{StGAN} & \textbf{TF} & \textbf{DuCoGAN}\\ 
\hline
Pororo & 4400 & \textbf{0.59} & 0.59 & 0.58 \\
Loopy & 2279 & 0.07 & 0.08 & \textbf{0.21} \\
Crong & 3327 & 0.50 & \textbf{0.51} & 0.49 \\
Eddy & 3154 & 0.48 & 0.50 & \textbf{0.58} \\
Poby & 2346 & 0.25 & 0.26 & \textbf{0.44} \\
Petty & 1564 & 0.16 & 0.17 & \textbf{0.49} \\
Tongtong & 717 & \textbf{0.15} & 0.16 & 0.14\\
Rody & 1073 & 0.21 & 0.20 & \textbf{0.41}\\
Harry & 1503 & 0.40 & 0.41 & \textbf{0.42}\\
\hline
\end{tabular}
\caption{\label{tab:char_scores} Character Classification F-Scores on Pororo-SV validation set (StGAN=StoryGAN, TF=StGAN+Transformer).}
\vspace{-10pt}
\end{table}

\paragraph{Individual Character Accuracy.}
As detailed in the Experiments section in the main paper, there are 9 characters which are featured throughout the Pororo-SV dataset. The distribution of characters varies across scenes, with some occurring more frequently than others. Using StoryGAN, Pororo, the most frequently occurring character in the dataset, has the highest F-Score, while the decrease in F-Score for other characters roughly correlates with their frequency in the data. With \modelname{}, we saw marginal improvements for Pororo and up to 30\% absolute improvement in F-Score for less frequent characters like Loopy. See Figure \ref{tab:char_scores} for a detailed breakdown of each character. While this confirms the data intensive nature of story visualization, it also shows that advanced modelling approaches can alleviate the issue of data scarcity to some extent. However, models in this domain will ultimately need to be extended to more diverse datasets with more characters and settings before they can be useful in practical applications (see Introduction).

\renewcommand{\arraystretch}{1.3}%
\begin{table}
\small
\centering
\begin{tabular}{|c|c|c|}
\hline
\textbf{Model} & \textbf{Metric} & Score \\ 
\hline
Inceptionv3 & Frame Acc. & 41.93 \\
 & Precision & 74.66 \\
  & Recall & 64.12 \\
 & F-Score & 68.99 \\
 & Accuracy & 80.68 \\
 \hline
 MART & BLEU2/3 & 38.05 / 33.59 \\
 & METEOR & 36.61 \\
 & ROUGE\_L & 44.98 \\
 \hline
H-DAMSM & R-Precision & 88.05 $\pm$ 0.00 \\

\hline
\end{tabular}
\caption{\label{tab:upper_bound} Upper Bounds of models used for Metrics on Pororo-SV validation set. }
\vspace{-10pt}
\end{table}

\paragraph{Evaluation Metric Upper Bounds.}

Many of the evaluation metrics that we use take advantage of other external model architectures (see Evaluation section in main paper), similar to prior work in this domain \cite{li2019storygan,LI2020102956}. Therefore, the quality of the evaluation metrics is contingent upon the accuracy of these models. Table \ref{tab:upper_bound} contains the upper bound results for these models on the Pororo-SV dataset. The finetuned Inceptionv3 model achieves high overall accuracy i.e. more than 85\% on validation and test sets. Video captioning model MART achieves high scores on the Pororo-SV validation set for several NLG metrics. The H-DAMSM model achieves 88.05\% R-precision on the validation set.

\paragraph{More Generated Examples.}
\begin{figure*}[t]
\centering
\includegraphics[width=0.70\textwidth]{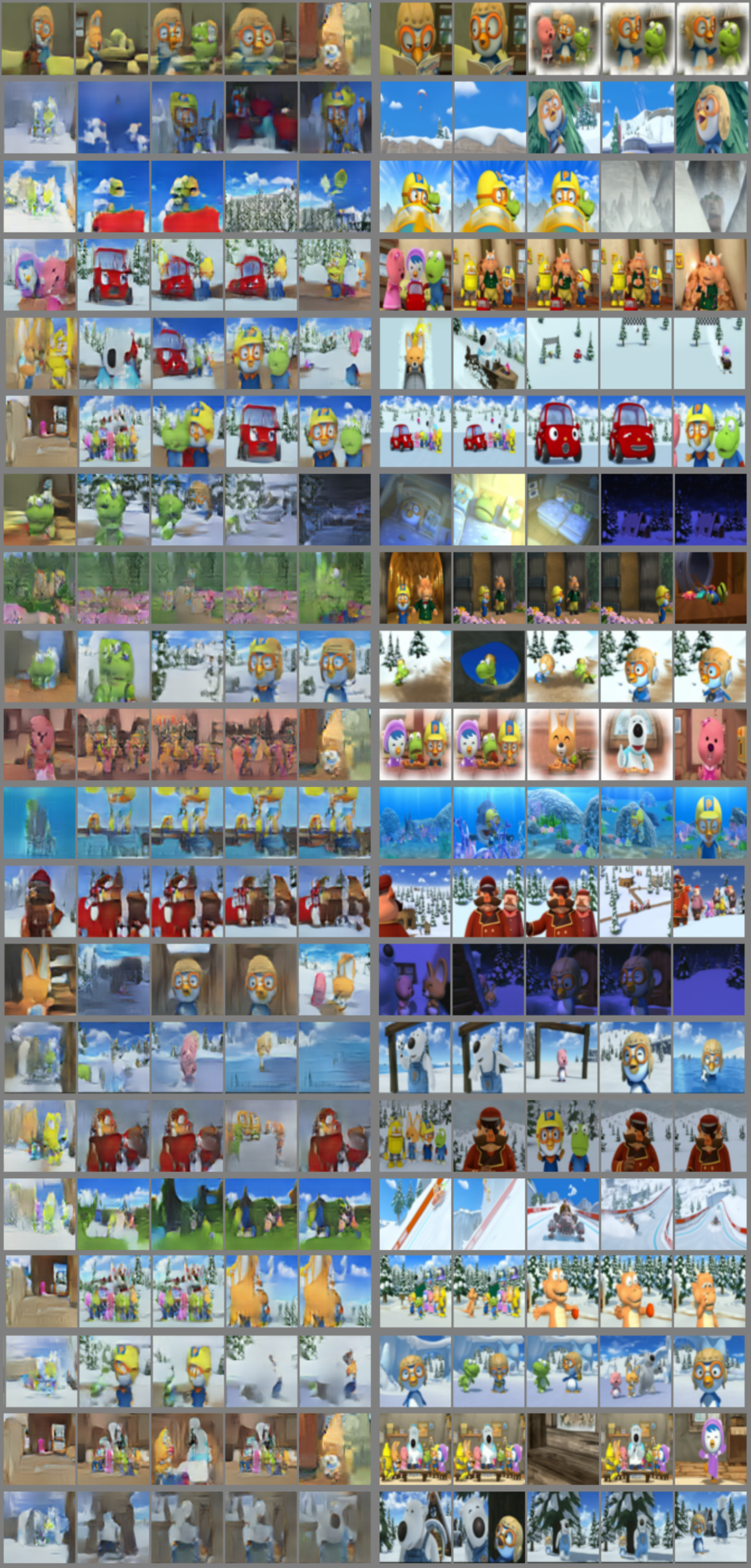}
\caption{Additional generated examples using our model. On the left is the generated examples and on the right is ground truth.
\label{fig:additional_examples}}
\end{figure*}
Figure \ref{fig:additional_examples} contains additional examples that our \modelname{} model generated. 

\end{document}